# Concept Discovery through Information Extraction in Restaurant Domain

Nadeesha Pathirana[1], Sandaru Seneviratne[1], Rangika Samarawickrama[1], Shane Wolff[1], Charith Chitraranjan[1], Uthayasanker Thayasivam[1], Tharindu Ranasinghe[2]

[1] Department of Computer Science and Engineering, Faculty of Engineering, University of Moratuwa, Sri Lanka.
[2] CodeGen International, Trace Expert City, Colombo 10, Sri Lanka.
{nadeesha.14, sandaru.14, rangika.14, shanewolff.14, charithc, rtuthaya, tharindu.10} @cse.mrt.ac.lk

**Abstract.** Concept identification is a crucial step in understanding and building a knowledge base for any particular domain. However, it is not a simple task in very large domains such as restaurants and hotel. In this paper, a novel approach of identifying a concept hierarchy and classifying unseen words into identified concepts related to restaurant domain is presented. Sorting, identifying, classifying of domain-related words manually is tedious and therefore, the proposed process is automated to a great extent. Word embedding, hierarchical clustering, classification algorithms are effectively used to obtain concepts related to the restaurant domain. Further, this approach can also be extended to create a semi-automatic ontology on restaurant domain.

## 1 Introduction

At present, there exists an astounding amount of data available which can be used to understand a specific domain along with the important building blocks of the domain. This data can further be used in the construction of a comprehensive ontology which can be interpreted by the relevant and important concepts of the domain. Considering the restaurant domain, to recognize, to understand, and to evaluate the concepts and relationships among them, a proper idea on how different aspects such as food, staff, atmosphere are spread across the domain is required. When the aspect food is taken into consideration, there are multiple sub categories such as, seafood, meat, desserts etc. These subcategories can be further divided forming hierarchies. Hence, the manual process of concept identification for the restaurant domain is problematic. Therefore, to avoid the hassle caused by the manual process of concept identification, research efforts have focused on semi-automatic process of concept identification for the restaurant domain.

Semi-automatic process of concept identification in a particular domain consists mainly of developing a computational model which enables capturing the meaning of the words along with the relative meaning and similarity among words in a text corpus to identify the aspects specific to the domain and concepts which the aspects belong to. This requirement can be modeled through a high dimensional vector space which provides a vector representation for each word in the corpus. This numerical vector representation can be used in computing the similarity measure between the pairs of words through which, the clustering of words into similar groups can be done. These clusters of similar words can be identified as different classes in the domain which represent important aspects specific to that domain. Thus, the concepts of the domain can be obtained. Further, the concepts can be ordered hierarchically so the procedure can be extended so as to construct an ontology as required.

This study proposes a novel way of obtaining the concepts and aspects in the restaurant domain, semi-automatically through which, the ontology can be built. In the proposed methodology, hierarchical clustering was performed for aspects in the restaurant domain using word embedding trained through Word2Vec model. Various approaches can be adopted in order to obtain the clusters from a hierarchical clustering. This research proposes a novel way of obtaining the optimal number of clusters from a hierarchical clustering using the silhouette index. The obtained clusters are then, refined manually to identify the concepts associated with the restaurant domain through which the restaurant-related entity classes are identified and these classes are available to use in the construction of the restaurant ontology.

The structure of the paper is as follows. The related work to this study is reviewed in section 2. Section 3 presents the methodology that we have adopted to develop the proposed process. Section 4 reports on the results obtained. Section 5 concludes the paper and suggests directions for future work.

## 2  Background and Related Work

### 2.1  Word Vector Embedding

First proposed by Tomas Mikolov, word vector embedding in Natural Language Processing (NLP) and feature learning is an approach of how words from a vocabulary are mapped to vectors considering the semantic meaning of words and relationship among words. This vector representation initially had a vector space with one dimension for each word. But due to the difficulties encountered with the number of dimensions, the vector space evolved to a lower dimension.

There is a number of word embedding models like Word2vec [1], GloVe [2], Latent semantic analysis (LSA) and Latent Dirichlet Allocation (LDA) which have their own advantages and drawbacks depending on the task the model is being used. LSA and LDA are popular models for statistical information related tasks while the use of these models on analogy tasks is poor [2].

Unlike LSA and LDA, Word2vec is considered to perform well in analogy tasks [2], due to its nature of considering the surrounding words and target words when creating the vectors for words. This model has two main models as Skip Gram [3] and Continuous Bag of Words (CBOW) [4]. In Skip Gram model, the context words are predicted using the target word whereas in CBOW model, the target word is predicted using the context words. GloVe model is also considered good in analogy tasks which uses a matrix factorization mechanism, combining the advantages of the Skip Gram model in Word2Vec [2] in predicting the vectors of words in text corpus. Hence, in this study, Word2Vec and GloVe models are used.

### 2.2 Clustering

Clustering is a popular field of study in data mining [5] which is used abundantly in statistical data analysis. Clustering can be defined as the mechanism of grouping a set of items, objects into clusters based on the characteristics in such a way that high intra class similarity and low inter class similarity features are preserved.

There are different algorithms which can be used in clustering data items. But depending on the task at hand, the algorithm used can vary. For this study, hierarchical clustering is used since the number of clusters/classes that are required for identifying the concepts is not defined beforehand. There are two hierarchical clustering algorithms; agglomerative (bottom up) [6] and divisive [7] (top down). In agglomerative approach, incrementally, two clusters which are most similar are merged and hierarchical clustering is performed which is represented through a tree like structure [8] named dendrogram. In divisive approach, clusters are recursively split creating the dendrogram. In this study, agglomerative approach is used to create the hierarchical clustering because it is less time consuming and efficient to merge clusters than divisive approach [9].

Given a dendrogram, obtaining the optimal number of clusters is a challenging task for we need to make sure the cluster quality is preserved with high intra class similarity and low inter class similarity. There are many indices that have been introduced by researchers to obtain the quality of clusters through which the number of clusters to be obtained from a dendrogram can be identified. This study incorporates the average silhouette index [10] to measure the cluster quality through which the optimal number of clusters in the hierarchical clustering can be obtained. In a previous research [11], Compact Separate Proportion (CSP) has been used for the same task.

### 2.3 Ontology Construction

Ontology can be identified as a formal representation of the concepts and relationships among the concepts in a specific domain. Manual construction of an ontology requires extensive efforts and expertise in a specific domain and is considered as an inherently complex task in the field of Natural Language and Processing. There are different approaches that have been adopted by researches in creating ontologies. One of the most common approaches is the conversion of database into an ontology [12]. Another approach is using rules to develop the ontology. These approaches have their own advantages and disadvantages. In this study, a method has been proposed to semi automate the process of identifying the concepts. It can be used to construct an ontology using word embedding and hierarchical clustering [13].

## 3  Methodology

This section discusses the methodology used for identifying the ontology entities. Each of the following subsections describe the steps of each process.

### 3.1  Data Collection & Data Preprocessing

To conduct this research a significant amount of restaurant domain related data is required as the objective is to build a more accurate and a detailed domain specific knowledge base. For that, user reviews are an ideal method to capture every aspect of restaurant domain like food, beverages, staff, environment etc. Therefore, around 1 million user reviews have been obtained from different regions and different countries.

Apart from the user reviews set, another two large text corpora have been used to conduct this research. One is Stanford 10-million-word set which is freely available for research purposes and the other one is w2v_gbg data set. For further applications, these three raw data sets were transformed into usable formats.

As the first step we eliminated all the numbers, emojis and other non-text data from the tokenized sentences from the data sets. Afterwards, stop words were eliminated since they carry less important meaning than keywords [14] and they are irrelevant for the restaurant domain. Next, lemmatization was done. The process of mapping inflected forms of a word into its base form or lemma is called lemmatization. Even after removing non-alphabetic characters and stop words from the data sets, having different forms of a word can affect the final result obtained by having different vectors for the words which are semantically equal. It also unnecessarily consumes space and affects the word embedding model. Therefore, the text corpora were lemmatized to obtain the lemma of the words. After that, all words were converted into lowercase in order to prevent duplicates of the same word with various cases.

### 3.2  Extracting Frequent Nouns

In the process of identification of the ontology classes for the restaurant domain, distinguishing the nouns that are related to the restaurant domain is vital. To obtain the domain related information, the nouns from the restaurant user reviews were used. First, 10 000 most frequent nouns from the review data set were obtained which can be identified as domain related keywords [15]. To make sure only the nouns are filtered out and to improve the accuracy, two different libraries were used. Next step was to filter the nouns whose cosine distances between them and candidate nouns were lesser than a threshold level. The candidate nouns were 'restaurant', 'food', and 'beverage' which can be identified as domain specific nouns. Through that process, around 1500 nouns from the restaurant user reviews dataset which are more relevant to the restaurant domain were obtained.

### 3.3 Training Word2Vec and GloVe models

In this stage we trained word embedding models for the collected three text corpora. Restaurant review text corpus contains reviews around 1 million whereas Stanford dataset contains text around 10 million texts and the w2v_gbg dataset contains 1 million texts.

Using GloVe algorithm, we built several GloVe models for the collected three text corpora. To build various GloVe models, different parameter values like dimension of the vector, window size, number of iterations, whether the context is symmetric or asymmetric etc. were changed. In Word2Vec, the parameters like dimension of the vector, window size, minimum frequency of a word, whether to use SG or CBOW architecture etc. were considered and several Word2Vec models were built [16, 17].

### 3.4 Hierarchical Clustering

To build a more comprehensive knowledge base for the restaurant domain, main ontology classes in the domain should be identified. Since constructing the hierarchy of concepts manually is difficult and can lead to missing out important concepts in the domain, hierarchical clustering can be used to identify the hierarchy and the clusters. The agglomerative hierarchical clustering [15] approach is adopted in this research.

To identify the domain specific concepts, around 1500 domain related keywords filtered from the 10000 most frequent nouns were used. When building hierarchical clustering, two parameters, linkage method and distance metric were used. Using different combinations of linkage method and distance metric for all the models that have been built from Word2Vec and GloVe, different hierarchical clustering were obtained and relevant dendrogram for each was obtained. It is a visualization of hierarchical relationships of the built hierarchical clustering. From all the hierarchical clustering that have been built from each model, the most suitable model for the research was identified using the highest cophenetic coefficient ($c$) calculated as in equation 1.

$$c = \frac{\sum_{i<j}(x(i,j)-\bar{x})(t(i,j)-\bar{t})}{\sqrt{[\sum_{i<j}(x(i,j)-\bar{x})^2][\sum_{i<j}(t(i,j)-\bar{t})^2]}} \quad (1)$$

The terms in equation 1 are defined as follows.

- $x(i,j) = |X_i - X_j|$; the ordinary Euclidean distance between the $i^{th}$ and $j^{th}$ observations
- $t(i,j)$; the dendrogrammatic distance between the model points $T_i$ and $T_j$
- $\bar{x}$; average of $x(i,j)$
- $\bar{t}$; average of $x(i,j)$

Cophenetic coefficient indicates how the word similarity compared to actual word similarity is preserved by the dendrogram and it is a linear correlation between the dissimilarity between each pair of observations and their corresponding cophenetic distances.

Comparing the cophenetic coefficient of each model, the most accurate model for the hierarchical clustering was identified with the following parameters.

- Algorithm:        Skip-gram
- Vector Size:      300
- Window Size:      5
- Minimum Count:    5

## 3.5 Identifying Clusters

After the identification of most suitable dendrogram for the research purpose, the next step was to discover the clusters within dendrogram to identify the concepts in the restaurant domain. Even though it is fairly easy to obtain clusters simply by setting a horizontal cut off line at a suitable distance, it does not guarantee that it would be the optimal number of clusters that could be obtained from the dendrogram while preserving the cluster quality. So a novel approach to measure the quality of clusters using intra-cluster similarity and inter cluster dissimilarity is used.

The silhouette index simply calculates the degree of cohesion of the objects within a cluster compared to the other clusters. The comparison attempts to capture the degree of separation. The index magnitude ranges from -1 to +1. Silhouette index can be calculated using any distance metric like Euclidean, Manhattan etc. Let *i* corresponds to a data point then *a(i)* is defined as the average distance from the data point to all other data points in the same cluster. The smallest average distance from *i* to all data points in any other cluster is defined as *b(i)*. Now the silhouette index can be formulated as mentioned in equation 2.

$$s(i) = \frac{b(i) - a(i)}{\max\{a(i), b(i)\}} \qquad (2)$$

Using silhouette index, the average index value for a dendrogram was calculated by considering overall cluster quality. From all the calculated average indices for each possible number of clusters in the dendrogram, the maximum index value was picked. Thus the optimal number of clusters which corresponds to the maximum average silhouette index was obtained.

### 3.6 Construction of Concept Hierarchy

Knowing the optimal number of clusters using silhouette index is a way of estimating how many leaf level concepts should be available for a given knowledge base. Since, the rough estimate of optimal number of clusters is available at this step, it is possible to construct the concept hierarchy for the knowledge base by suitably investigating the clusters. Given the rough clusters and the dendrogram of the words, identification of the concepts requires human intervention. It is not impossible to automate this process but the accuracy of identified concepts from the clusters may vastly deviate from what is actually expected. This is due to the fact that any given cluster can be conceptualized (the process of identifying the common characteristics of the words in the cluster and identifying a name which represent the group of words the best) in various ways in relation to the aspect we look into it. E.g. the cluster of words {spoon, fork, knife} can be conceptualized differently as cutleries, tableware etc.

The dendrogram reveals how clusters are hierarchically merged. Therefore, the rough clusters are manually checked against the dendrogram and the concept names are decided suitably for each cluster. In this process, some branches of the dendrogram appeared to be merged inappropriately. Such branches are excluded as and when necessary to preserve the cluster quality and the coherence of the concepts. Finalizing the leaf level concepts enabled the rest of the work which was to identify the next level concepts up in the hierarchy. While preserving the merging structure of the original dendrogram, the merged nodes were suitably conceptualized until the top root is arrived. At the end of the conceptualization, the concept hierarchy was obtained.

### 3.7 Classification

After identifying the concepts, it is required to classify any previously unseen words into the correct cluster/concept so that the classifier can be used to populate the ontology created from the concept hierarchy for totally unseen words. E.g. the cluster tableware includes all types of dishes and cutleries. When an unseen word like 'mug' is encountered, the classifier is expected to classify it under the tableware cluster. In order to classify the words as required, different classification algorithms were used. An artificial neural network classifier, k – nearest neighbor classifier and random forest classifier were tested with acquired cluster data.

## 4 Results

Three text corpora have been used to build Word2Vec and Glove models and dendrograms for each model have been obtained. When observing the models created from the Stanford-10-million-word set, we could identify some words which did not belong to the clusters they were in with respect to the restaurant domain. For an example, the word 'apple' was modeled as a word related to mobile, WIFI, GSM whereas in the restaurant domain, the word 'apple' belongs to the fruit category. The reason was, in the dataset, the word 'apple' was identified as a brand name rather than a fruit. When

w2v_gbgenl dataset was used, the results improved comparatively to Stanford dataset, but we could identify words which were clustered together as food, but not in specific clusters like, fruits, dessert, fish etc. The reason was, the dataset contained very general information which did not contain much information about the restaurant domain. Hence, when the model was created using this dataset, there was not enough information to model the vectors in such a way that clear clusters are formed. The restaurant review dataset gave further improved and more accurate results than both the Stanford and w2v_gbgenl datasets as it contains more domain specific information and hence clear clusters in the dendrogram could be identified.

The models were created using the cosine distance metric as semantic similarity is considered in cosine distance unlike in the Euclidean distance. Moreover, when the cosine distance metric is used, the best method to generate the dendrogram is using the average linkage method [18].

Table 1 shows the specifications of the Word2Vec models created using the cosine distance metric, average linkage method, windows size = 5, min count = 5 and vector size = 300 along with the cophenetic coefficients obtained. Closer the cophenetic coefficient to 1, better the model is.

**Table 1.** Word2vec Model Results

| Dataset | Model No. | Architecture | Cophenetic Coefficient |
| --- | --- | --- | --- |
| Stanford | W1 | Skip Gram | 0.494 |
| Stanford | W2 | CBOW | 0.398 |
| w2v_gbgenl | W3 | Skip Gram | 0.495 |
| w2v_gbgenl | W4 | CBOW | 0.483 |
| Restaurant reviews | W5 | Skip Gram | 0.542 |
| Restaurant reviews | W6 | CBOW | 0.491 |

Table 2 shows the Cophenetic coefficient results for GloVe models created with the windows size = 10, min count = 5, vector size = 300 and number of iterations = 15 for models created using cosine distance metric and average linkage method.

**Table 2.** GloVe Model Results

| Dataset | Model No. | Context | Cophenetic Coefficient |
| --- | --- | --- | --- |
| Stanford | G1 | Symmetric | 0.300 |
| Stanford | G2 | Asymmetric | 0.368 |
| w2v_gbgenl | G3 | Symmetric | 0.401 |
| w2v_gbgenl | G4 | Asymmetric | 0.460 |

| Restaurant reviews | G5 | Symmetric | 0.439 |
| Restaurant reviews | G6 | Asymmetric | 0.497 |

Based on the results obtained, the most accurate model to perform hierarchical clustering was identified as W5 model. Using this model, the dendrogram for the most frequent nouns was created.

In order to obtain the optimal number of clusters, silhouette index was calculated. The maximum index value 0.35 was recorded for 96 clusters. Apparently 0.35 score is low. This is due to the fact that cluster separation is low. Even if the cluster cohesion is high, silhouette index tends to decrease when inter cluster dissimilarity drops. Since we have filtered out all the words to fit the restaurant domain, the clusters appear closely related to each other. Therefore, silhouette index has become low as expected.

In the dendrogram there are clearly separated clusters which represent various concepts related to restaurant domain. Alcoholic beverages, types of coffee, types of fish and Indian cuisine (see Fig. 1 – 4 respectively) are some identified concepts.

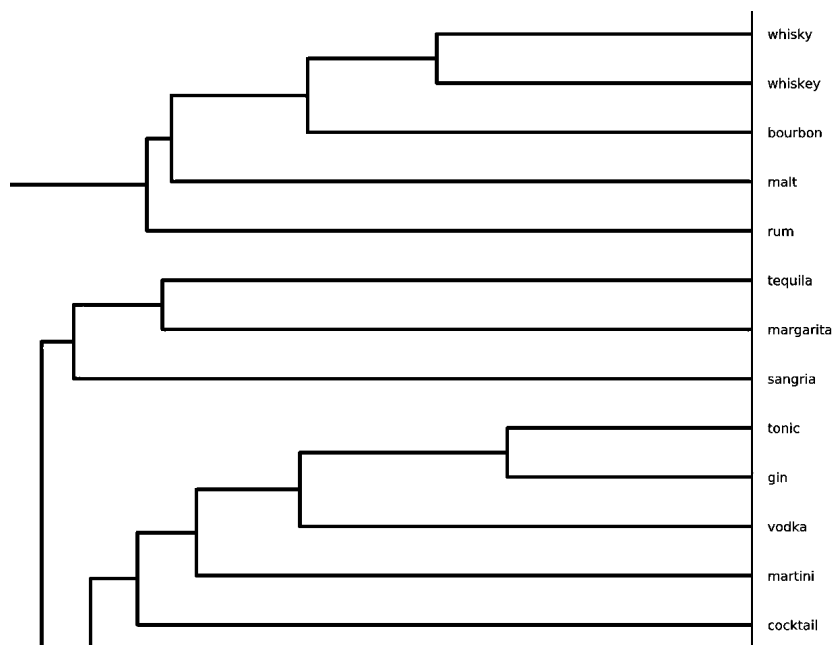

**Fig. 1.** Alcoholic Beverages

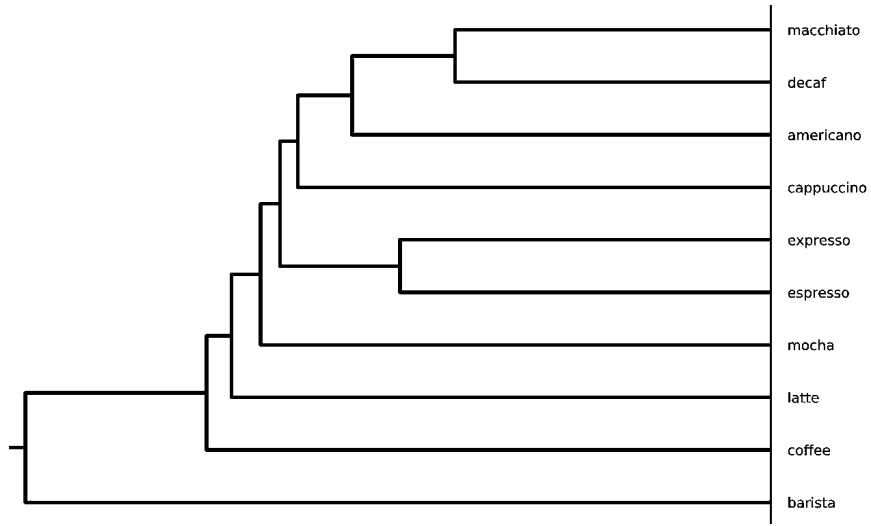

**Fig. 2.** Types of Coffee

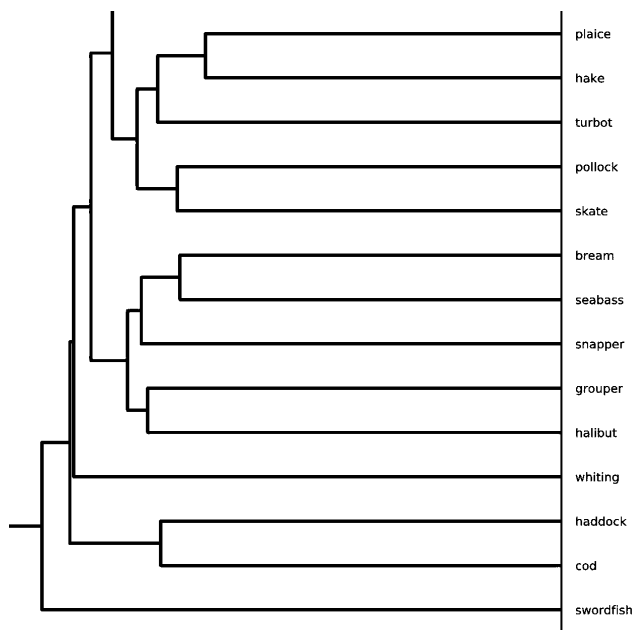

**Fig. 3.** Types of Fish

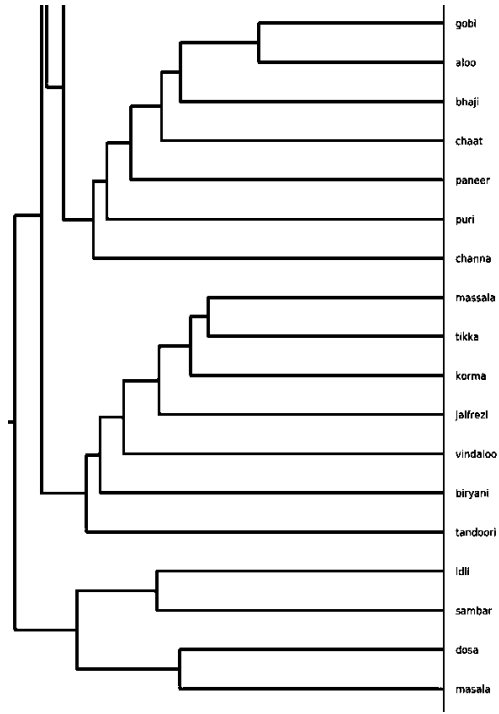

**Fig. 4.** Indian Cuisine

After obtaining the dendrogram, some of the clusters were manually refined to improve the quality and the accuracy further.

Three types of classifiers have been tested with the refined cluster data obtained. Out of them artificial neural network classifier outperformed the rest. The accuracy, precision, recall and F1 scores of the classifiers are given in table 3.

**Table 3.** Classification Results

| Classifier | Accuracy | Precision | Recall | F1 Score |
| --- | --- | --- | --- | --- |
| Artificial Neural Network (*Two hidden layers of 300 nodes each*) | 0.70 | 0.67 | 0.67 | 0.64 |
| K Nearest Neighbor (*k = 5*) | 0.30 | 0.19 | 0.30 | 0.21 |
| Random Forest (*100 estimators*) | 0.46 | 0.39 | 0.47 | 0.38 |

## 5     Conclusion and Future Works

This paper discusses concept identification for restaurant domain with the assistance of hierarchical clustering. In order to perform hierarchical clustering, the words should be vectorized. Word embedding models have been used to represent words as vectors. Then, the word-vector space was clustered using hierarchical clustering. In this novel approach, the importance of automating the process of identification of concepts was outlined. After estimating the optimal number of clusters through evaluating the average silhouette index, the classification of previously unseen words into the obtained clusters was performed. The accuracy of the classifier is vital since it leads to the correct identification of concepts embedded in a restaurant review using the concept hierarchy obtained. The proposed methodology can also be extended to other domains without restrictions, since the process is general irrespective of the domain. Converting the process to fit another domain is a matter of fine tuning the model to incorporate and domain-related concepts to best suit the aspects considered. As future work, this methodology can be extended such that obtained concepts are used in ontology building through which a comprehensive knowledge base for the restaurant domain can be acquired.